\documentclass{IEEEoj-data}
\usepackage{cite}
\usepackage{amsmath,amssymb,amsfonts}
\usepackage{algorithmic}
\usepackage{graphicx,color}
\usepackage{textcomp}
\usepackage{hyperref}
\usepackage{balance}
\usepackage{dirtree}
\usepackage{listings}
\usepackage{xcolor} 
\usepackage{hyperref}
\usepackage[ruled,vlined]{algorithm2e}
\def\BibTeX{{\rm B\kern-.05em{\sc i\kern-.025em b}\kern-.08em
    T\kern-.1667em\lower.7ex\hbox{E}\kern-.125emX}}
\AtBeginDocument{\definecolor{ojcolor}{cmyk}{0.93,0.59,0.15,0.02}}

\usepackage[acronym,toc]{glossaries}
\usepackage{booktabs}
\usepackage{graphicx,color}
\usepackage[table]{xcolor}
\definecolor{lightgray}{gray}{0.9}
\definecolor{darkgray}{gray}{0.7}

\glsdisablehyper

\newacronym{vlm}{VLM}{Vision-Language Model}
\newacronym{dtpqa}{DTPQA}{Distance-Annotated Traffic Perception Question Answering}
\newacronym{vqa}{VQA}{Visual Question Answering}
\newacronym{sota}{SOTA}{state-of-the-art}
\newacronym{llm}{LLM}{Large Language Model}
\newacronym{moe}{MoE}{Mixture of Experts}

\glsaddall

\begin{document}

\title{\textcolor{black}{Descriptor:} \textcolor{ieeedata}{\textit{Distance-Annotated Traffic Perception Question Answering (DTPQA)}}}

\author{NIKOS~THEODORIDIS$^{1,2,3}$ (Graduate Student Member, IEEE), TIM~BROPHY$^{1,2,3}$ (Member, IEEE), REENU~MOHANDAS$^{1,2,3}$ (Member, IEEE),  GANESH~SISTU$^{1,2,4}$, FIACHRA~COLLINS$^{4}$, ANTHONY~SCANLAN$^{1,2}$, CIARÁN~EISING$^{1,2,3}$ (Senior Member, IEEE)}
\affil{Department of Electronic and Computer Engineering, University of Limerick, Castletroy, Co. Limerick V94 T9PX, Ireland}
\affil{Data Driven Computer Engineering Research Centre, University of Limerick, Castletroy, Co. Limerick V94 T9PX, Ireland}
\affil{Lero, The Irish Software Research Centre, University of Limerick, Limerick V94 NYD3, Ireland}
\affil{Valeo Vision Systems, Dunmore Road, Tuam, Co. Galway H54 Y276, Ireland}
\corresp{CORRESPONDING AUTHOR: NIKOS THEODORIDIS (e-mail: theodoridis.nikolaos@ul.ie).}
\authornote{This work was supported in part by Science Foundation Ireland under Grant 13/RC/2094\_P2 and co-funded under the European Regional Development Fund
through the Southern and Eastern Regional Operational Programme to Lero - the Science Foundation Ireland Research Centre for Software, and in part by Valeo
Vision Systems.}
\markboth{Descriptor: Distance-Annotated Traffic Perception Question Answering (DTPQA)}{Theodoridis \textit{et al.}}

\begin{abstract}
The remarkable progress of \glspl{vlm} on a variety of tasks has raised interest in their application to automated driving. However, for these models to be trusted in such a safety-critical domain, they must first possess robust perception capabilities, i.e., they must be capable of understanding a traffic scene, which can often be highly complex, with many things happening simultaneously. Moreover, since critical objects and agents in traffic scenes are often at long distances, we require systems with not only strong perception capabilities at close distances (up to 20 meters), but also at long (30+ meters) range. Therefore, it is important to evaluate the perception capabilities of these models in isolation from other skills like reasoning or advanced world knowledge. \gls{dtpqa} is a \gls{vqa} benchmark designed specifically for this purpose: it can be used to evaluate the perception systems of \glspl{vlm} in traffic scenarios using trivial yet crucial questions relevant to driving decisions. It consists of two parts: a synthetic benchmark (DTP-Synthetic) created using a simulator, and a real-world benchmark (DTP-Real) built on top of existing images of real traffic scenes. Additionally, \gls{dtpqa} includes distance annotations, i.e., how far the object in question is from the camera. More specifically, each \gls{dtpqa} sample consists of (at least): (a) an image, (b) a question, (c) the ground truth answer, and (d) the distance of the object in question, enabling analysis of how \gls{vlm} performance degrades with increasing object distance. In this article, we provide the dataset itself along with the Python scripts used to create it, which can be used to generate additional data of the same kind.
 \\ 
 
 {\textcolor{ieeedata}{\abstractheadfont\bfseries{IEEE SOCIETY/COUNCIL}}}     IEEE Intelligent Transportation Systems Society (ITSS)\\  
 \\
 {\textcolor{ieeedata}{\abstractheadfont\bfseries{DATA DOI/PID}}}     \href{https://data.mendeley.com/datasets/9rj4kyrx9k/2}{10.17632/9rj4kyrx9k.2}\\
  
 {\textcolor{ieeedata}{\abstractheadfont\bfseries{DATA TYPE/LOCATION}}}  Images, .jpg files, .json files, University of Limerick, Ireland

\end{abstract}


\maketitle

\section*{BACKGROUND}
\label{background}
Despite the growing interest in applying \glspl{vlm} to automated driving, most existing benchmarks lack a focused evaluation of perception capabilities. We argue that a benchmark for evaluating the perception capabilities of \glspl{vlm} in traffic scenes, isolated from other skills, should meet at least three main criteria: (a) it must include trivial, perception-only visual questions that do not require any reasoning, (b) all samples should pertain to traffic scenes, with questions that are crucial for driving decisions (e.g., not questions like ``What colour is the car?''), and (c) it should use a multiple-choice format, eliminating the need for language-based evaluation metrics that can be affected by the model’s language generation abilities, which are not the focus here. Additionally, a useful extra feature is the inclusion of distance annotations, allowing the evaluation of \glspl{vlm}' perception capabilities as a function of the distance of the object in question, something important in traffic scenes, which include multiple agents and objects at varying distances. Many \gls{vqa} datasets for traffic scenes have been proposed (see Table \ref{traffic_vqa_overview}), but none of them meet all the above criteria. This is the gap we aim to fill with \gls{dtpqa}.

\begin{table*}[t]
    \centering
    \caption{Overview of VQA datasets for automated driving}
    \label{traffic_vqa_overview}
    \resizebox{\textwidth}{!}{
    \begin{tabular}{lcccc}
        \toprule
        Dataset & Trivial Perception & Driving-Relevant Questions & Distance Annotations & Multiple-choice \\
        \midrule
        LingoQA \cite{Marcu2024} & $\sim$ & $\checkmark$ & $\times$ & $\times$ \\
        DriveBench \cite{Xie2025} & $\times$ & $\checkmark$ & $\times$ & $\sim$ \\
        RoadTextVQA \cite{Tom2023} & $\checkmark$ & $\sim$ & $\times$ & $\checkmark$ \\
        SURDS \cite{guo2024surds} & $\sim$ & $\checkmark$ & $\sim$ & $\checkmark$ \\
        NuInstruct \cite{Ding2024} & $\times$ & $\checkmark$ & $\sim$ & $\checkmark$ \\
        TB-Bench \cite{Charoenpitaks2025} & $\times$ & $\checkmark$ & $\sim$ & $\checkmark$ \\
        STRIDE-QA \cite{ishihara2025stride} & $\times$ & $\checkmark$ & $\sim$ & $\checkmark$  \\
        TUMTraffic-VideoQA \cite{zhou2025tumtraffic} & $\sim$ & $\checkmark$ & $\times$ & $\sim$  \\
        \midrule
        DTPQA (ours) & $\checkmark$ & $\checkmark$ & $\checkmark$ & $\checkmark$ \\
        \bottomrule
        \multicolumn{5}{l}{$\checkmark$: feature present, $\times$: feature absent, $\sim$: partially present.}
    \end{tabular}%
    }
\end{table*}

\begin{figure*}[t]
    \centerline{\includegraphics[width=5in]{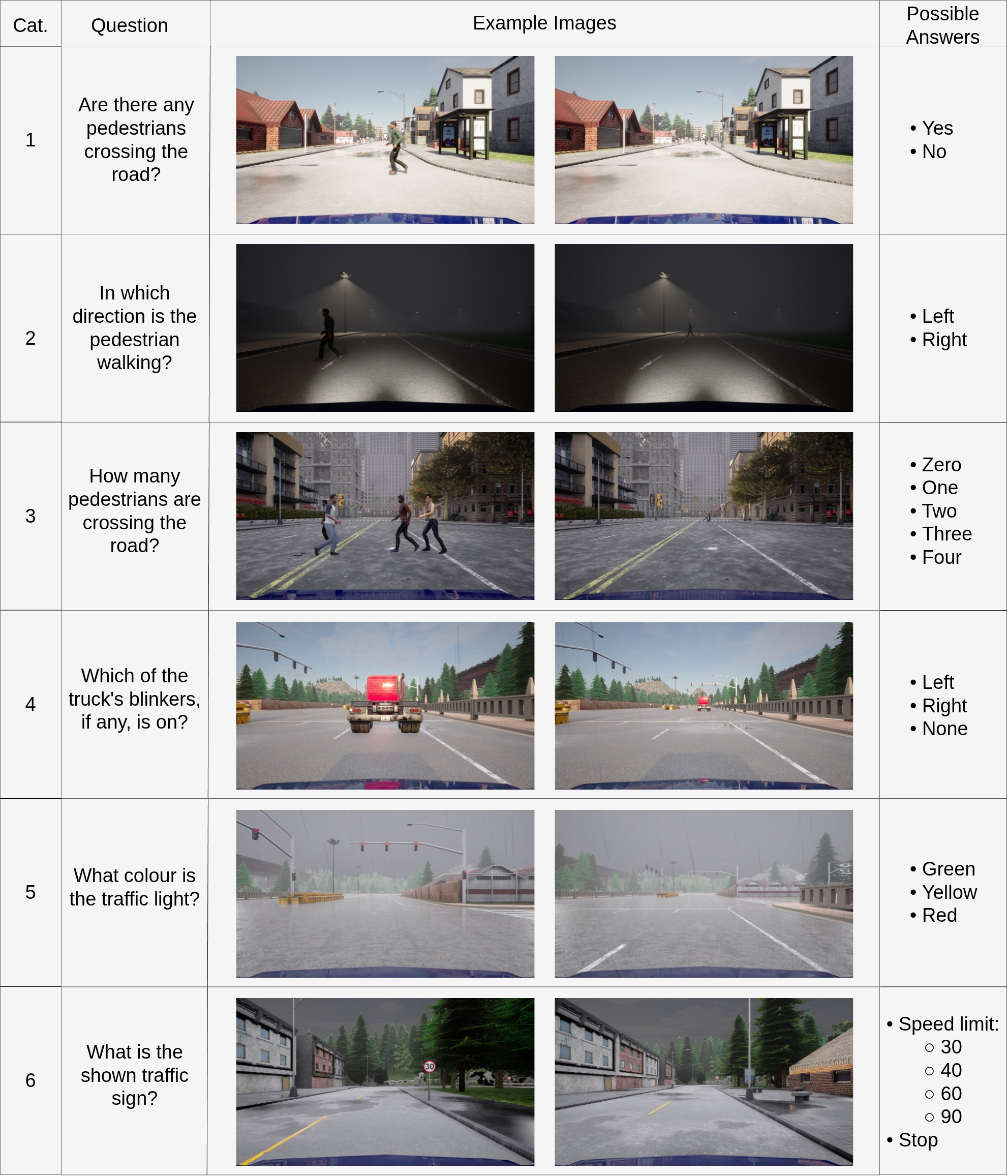}}
    \caption{The six different categories of samples in DTP-Synthetic. \label{DTP_synthetic_samples}}
\end{figure*}

\begin{figure*}
    \centerline{\includegraphics[width=5in]{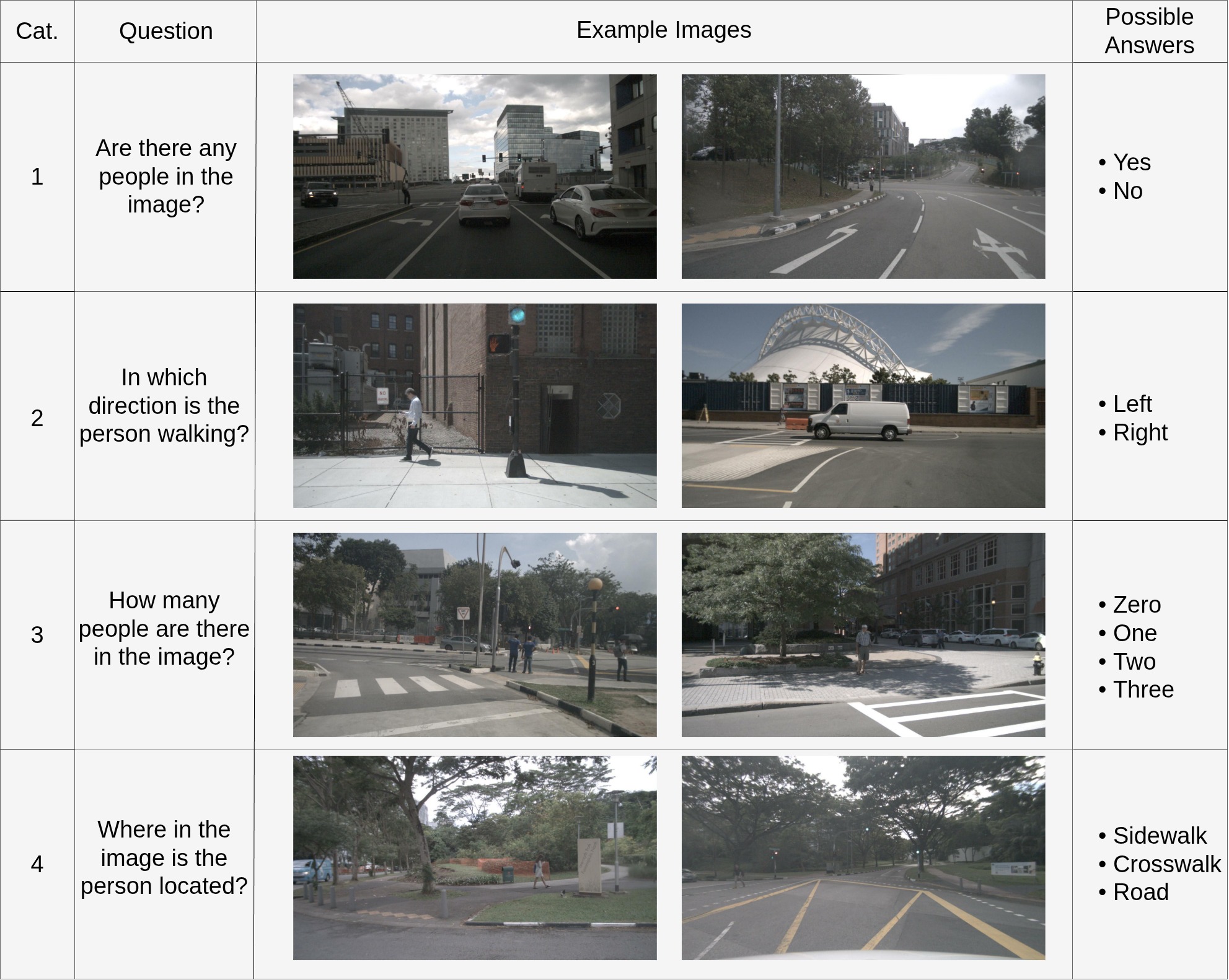}}
    \caption{The four different categories of samples in DTP-Real. \label{DTP_real_samples}}
\end{figure*}

\gls{dtpqa} consists of 19,149 samples, split into two parts: DTP-Synthetic (9,368 samples), created using the CARLA simulator \cite{Dosovitskiy2017}, and DTP-Real (9,781 samples), built on top of nuScenes \cite{Caesar2020}. \gls{dtpqa} includes variations of the same question across different object distances, ranging from 5 to 50 meters, with intermediate levels at 10, 20, 30, and 40 meters. It also includes cases where the object is entirely absent, referred to as \textit{negative samples}. The specific number of samples per category and per distance can be found in Table \ref{dtp_detailed_stats}. DTP-Synthetic contains six different categories of samples, and DTP-Real contains four, as shown in Figures \ref{DTP_synthetic_samples} and \ref{DTP_real_samples}, respectively. These categories were selected to satisfy two main criteria: (a) the question must concern a simple visual concept, and (b) it must be relevant to driving decisions. For instance, knowing the direction of a walking pedestrian (category 2 in Figures \ref{DTP_synthetic_samples} and \ref{DTP_real_samples}) is crucial when planning upcoming driving actions. For example, whether a pedestrian crossing the road is moving left or right can determine whether the driver should yield or proceed, depending on the pedestrian’s position. We also included both coarse visual questions, such as the presence of a pedestrian in the scene (category 1 in Figures \ref{DTP_synthetic_samples} and \ref{DTP_real_samples}), and fine-grained ones, such as the position (left or right) of the active blinker on the vehicle ahead (category 4 in Figure \ref{DTP_synthetic_samples}). \gls{dtpqa} was generated entirely on a machine equipped with an Intel Core i9-14900F CPU, an NVIDIA GeForce RTX 4070 Ti SUPER GPU with 16 GB of VRAM, and 32 GB of system RAM.

A key feature of \gls{dtpqa} is that it maintains a balanced number of samples for each possible answer at every distance. As shown in Table \ref{dtp_detailed_stats}, the number of samples at each distance within a category is always divisible by the number of possible answers for that category. For example, in Table \ref{dtp_detailed_stats}, there are 396 samples involving people on either a sidewalk, crosswalk, or the road (Cat.4-Real) at a distance of 30 meters in DTP-Real. Since there are three possible answers, the 396 samples are divided equally: 132 samples with a pedestrian on a sidewalk, 132 on a crosswalk, and 132 on the road, all at 30 meters. The same holds for all categories and distances. The only exception is Cat.5-Synth at 10 meters, where there are 12 samples with the answer ``Red'' and 11 each for ``Yellow'' and ``Green''\footnote{We consider this slight discrepancy to be statistically insignificant and to have no meaningful impact on the evaluation of model performance for this category.}. This balance is crucial, as it prevents models from gaining an advantage based solely on language biases. For categories 1 and 3, this divisibility rule applies to the number of positive answers only, excluding the negative cases, as these are not associated with a specific distance.

\begin{table*}[t]
    \centering
    \caption{Number of samples per category per distance}
    \label{dtp_detailed_stats}
    \resizebox{\textwidth}{!}{
    \begin{tabular}{l *{7}{c}>{\columncolor{lightgray}}c *{7}{c}>{\columncolor{lightgray}}c}
        \toprule
        \multicolumn{1}{l}{} & \multicolumn{8}{c}{\textbf{DTP-Synthetic}} & \multicolumn{8}{c}{\textbf{DTP-Real}} \\
        \cmidrule(lr){2-9}
        \cmidrule(lr){10-17}
        \textbf{Distances} & \textbf{5 m} & \textbf{10 m} & \textbf{20 m} & \textbf{30 m} & \textbf{40 m} & \textbf{50 m} & \textbf{Negative} & \textbf{Total} & \textbf{5 m} & \textbf{10 m} & \textbf{20 m} & \textbf{30 m} & \textbf{40 m} & \textbf{50 m} & \textbf{Negative} & \textbf{Total} \\
        \midrule
        \textbf{Category 1} & 180 & 180 & 180 & 180 & 180 & 180 & 180 & 1260 & 31 & 303 & 495 & 660 & 491 & 401 & 200 & 2581 \\
        \midrule
        \textbf{Category 2} & 180 & 180 & 180 & 180 & 180 & 180 & - & 1080 & 156 & 580 & 334 & 198 & 102 & 62 & - & 1432 \\
        \midrule
        \textbf{Category 3} & 720 & 720 & 720 & 720 & 720 & 720 & 180 & 4500 & 45 & 651 & 1053 & 1056 & 432 & 162 & 200 & 3599 \\
        \midrule
        \textbf{Category 4} & 270 & 270 & 270 & 270 & 270 & 270 & - & 1620 & 147 & 684 & 447 & 396 & 255 & 240 & - & 2169 \\
        \midrule
        \textbf{Category 5} & - & 34 & 156 & 156 & 156 & 156 & - & 658 & - & - & - & - & - & - & - & - \\
        \midrule
        \textbf{Category 6} & - & 50 & 50 & 50 & 50 & 50 & - & 250 & - & - & - & - & - & - & - & - \\
        \midrule
        \rowcolor{lightgray}\textbf{Total} & 1350 & 1434 & 1556 & 1556 & 1556 & 1556 & 360 & \cellcolor{gray}9368 & 379 & 1280 & 865 & 2329 & 2310 & 2218 & 400 & \cellcolor{gray}9781 \\
        \bottomrule
    \end{tabular}%
    }
\end{table*}

\gls{dtpqa} has been used to evaluate the perception capabilities of \gls{sota} small \glspl{vlm} in traffic scenes \cite{11230063} and it could be easily reused for evaluating any \gls{vlm}.

\section*{COLLECTION METHODS AND DESIGN}
\label{collection_methods}
In the subsections below, we describe the methodologies used for creating the data in DTP-Synthetic and DTP-Real, as well as the final processing steps that led to the final form of the dataset.

\subsection*{DTP-Synthetic}
As mentioned earlier, we created DTP-Synthetic using the CARLA simulator \cite{Dosovitskiy2017}. CARLA offers several advantages for creating a traffic VQA dataset focused solely on perception capabilities, such as the flexibility to customise the traffic scene exactly as desired, and the ability to generate identical scenes on demand while varying only one thing at a time, for example, the distance of the object/agent in question from the camera.

\begin{figure*}    
\centerline{\includegraphics[width=5in]{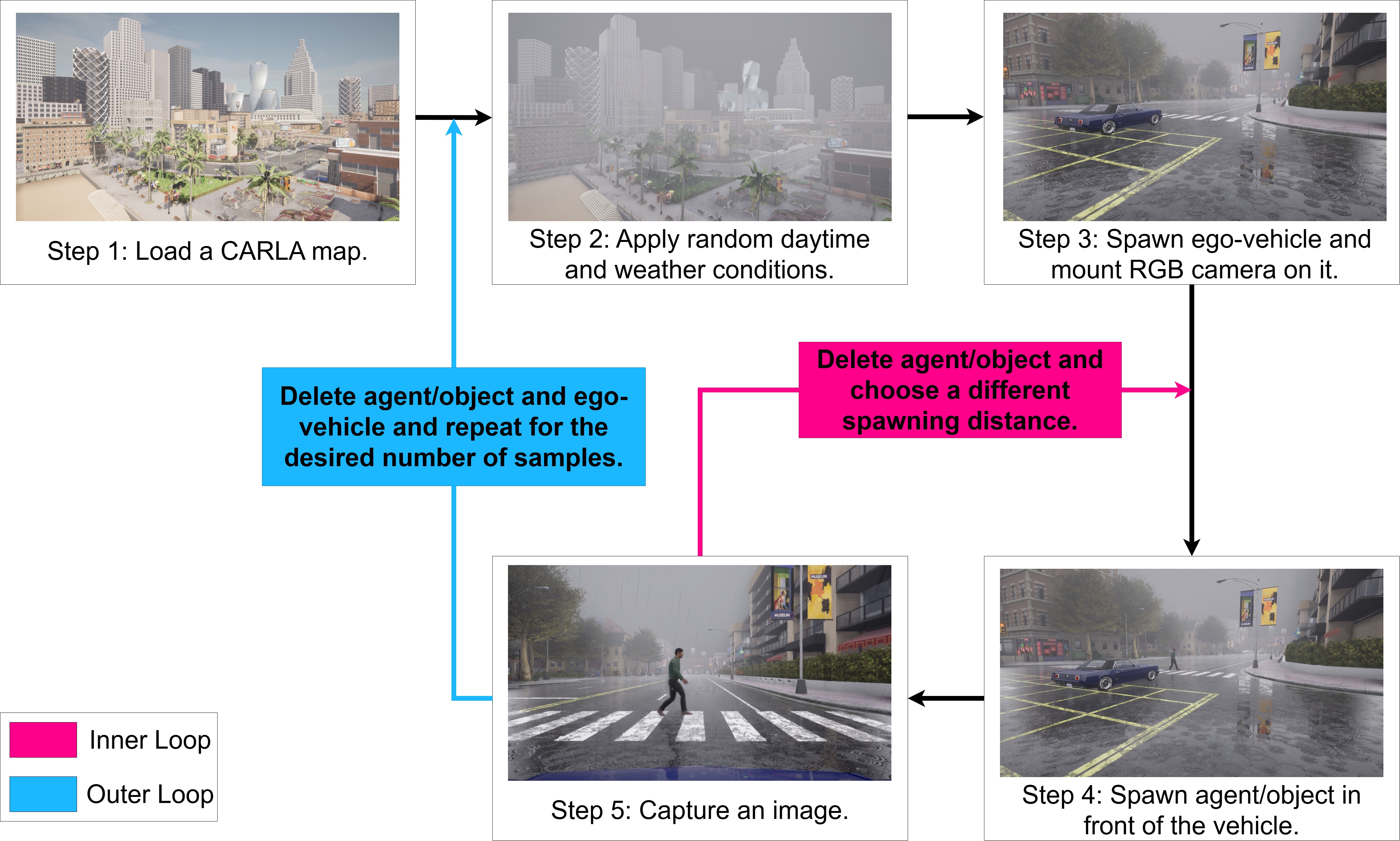}}
    \caption{The creation pipeline of DTP-Synthetic. \label{dtp_synthetic_creation_diagram}}
\end{figure*}

While certain details differ between the creation pipelines for each sample category, the high-level procedure remains the same and is depicted in Figure \ref{dtp_synthetic_creation_diagram}. In short, we start by loading a CARLA map, applying random daytime and weather conditions, spawning the ego-vehicle at a random (valid) point, and mounting an RGB camera on it. We then spawn the target agent/object at the desired distance and capture an image with the RGB camera, saving it together with the corresponding annotations (filename, question, answer, distance, town, weather, category-specific information). Next, we delete the agent/object from the map, respawn it at a different distance, and repeat the process. Once all desired distances have been covered, we delete all spawned objects (i.e., the ego-vehicle, RGB camera, and the agent/object) and repeat the procedure from step 2 in Figure \ref{dtp_synthetic_creation_diagram} for the desired number of samples in the loaded map. This entire process was repeated for multiple maps: Town01–07, Town10HD, Town12, and Town15. Below, we provide the specific details for creating the data in each category of DTP-Synthetic.

\subsubsection*{Categories 1 and 2 (Pedestrian ahead)}
The data for Cat.1-Synth and Cat.2-Synth were created in parallel, as they depict the same type of traffic scene (i.e., a single pedestrian crossing the road) with only the question changing. The only additional detail beyond the pipeline in Figure \ref{dtp_synthetic_creation_diagram} is that the pedestrian’s direction was chosen randomly. We always checked whether the pedestrian was spawned at a valid location (i.e., road or shoulder) so that the question “Are there any pedestrians \textbf{crossing the road}?” was meaningful. If this was not the case, we deleted all spawned objects and restarted.

\subsubsection*{Category 3 (Multiple pedestrians ahead)}
Cat.3-Synth follows the same procedure as Categories 1 and 2, except that we repeated the process four times per distance, first spawning one pedestrian, then two, and so on. We also added small Gaussian noise ($std=1.2$) to each pedestrian’s distance from the vehicle so that they would not all appear in exactly the same location.

\subsubsection*{Category 4 (Truck ahead)}
In Cat.4-Synth, we aim to evaluate whether \glspl{vlm} can identify which indicator is on in the vehicle ahead as a function of distance. To do this, we followed the pipeline in Figure \ref{dtp_synthetic_creation_diagram}, but spawned a truck\footnote{We chose a truck because most vehicles in CARLA have very small indicator lights that are difficult to spot at greater distances. The truck we used was an exception.} instead of a pedestrian, and randomly activated either its right indicator, left indicator, or none.

A challenge here was that indicator lights blink and are not always on, meaning the RGB camera might capture an image when the indicator was off. To address this, we captured multiple images per scene (some of which would capture the indicator when lit) and implemented a function to compare the brightness of the target indicator’s region with the opposite side (compare the indicator that is supposed to be on with the indicator that is supposed to be off). If the target side was brighter, the image was marked as a successful capture. From these “draft” images, we selected the one with the highest brightness in the target indicator region for inclusion in the dataset.

\subsubsection*{Category 5 (Traffic light ahead)}
Cat.5-Synth includes samples of traffic lights at varying distances, with questions about their colour. Here, the key difference from the high-level pipeline is that the object in question (the traffic light) was not deleted and respawned, as its position is fixed, so instead, the vehicle and mounted camera were moved. Since the vehicle must be on the road and within a driving lane, we could not simply place it in a straight line from the traffic light (e.g., in the case of curvy roads). Algorithm \ref{algorithm_1} shows how we correctly located the ego-vehicle in the map.

\begin{algorithm}
    \caption{Spawn vehicle at the correct position for Cat.5-Synth}
    \label{algorithm_1}
    For distance $x$:
    \begin{enumerate}
        \item Starting from the traffic light, follow a straight line \\ in the direction it faces to find point $x_1$ located at distance $x$.
        \item Retrieve all driving-lane waypoints within a radius \\ of 5 meters from $x_1$.
        \item Keep only the waypoints whose distance from the traffic light is $x \pm 2$ meters.
        \item From these candidates, select the waypoint whose orientation is closest to the target orientation (i.e., facing the traffic light), allowing for a maximum difference of 25 degrees.
        \item If such a waypoint exists, spawn the vehicle and camera at this location. Otherwise, skip this traffic light and proceed to the next, as it is unsuitable \\ for sampling (e.g., due to excessive road \\ curvature).
    \end{enumerate}
\end{algorithm}

It is worth noting that we were unable to capture samples at 5 meters because most traffic lights are elevated and fall outside the camera’s field of view at such a short range. The same applies at 10 meters for all towns except Town01 and Town02, where traffic lights are lower (hence the smaller number of samples at this distance for Cat.5-Synth in Table \ref{dtp_detailed_stats}).

\subsubsection*{Category 6 (Traffic sign ahead)}
The process for Cat.6-Synth is the same as for Cat.5-Synth, as the objects in question (traffic signs) are also fixed in the map. The only additional step was removing certain traffic signs that were problematic because they were too close to other traffic signs. For example, let's say our object in question is a speed limit sign (e.g., 30 km/h) located 50 meters from the camera, but there is also a stop sign at 10 meters from the camera. Then when asking the question \textit{``What is the shown traffic sign?"}, we can't expect any model (or even human) to notice the distant speed limit sign. In such cases, we removed from the map all but one of the traffic signs that were in close proximity.

\subsubsection*{Final Data Validation}
When generating thousands of samples in simulators like CARLA, errors are inevitable. Therefore, the final step in creating the synthetic data was to manually review all simulated images and remove any problematic samples. The main types of problematic samples that were removed fall into one of the following categories:

\begin{itemize}
    \item \textbf{Incorrect or failed spawn location}: The object of interest spawned at an incorrect location or did not spawn at all.
    \item \textbf{Occlusion}: The object of interest was not visible due to road geometry. For example, a pedestrian could spawn at a lower altitude than the vehicle because of a downhill road segment, resulting in the pedestrian being completely outside the camera's field of view.
    \item \textbf{Simulator rendering glitches}: Texture artefacts, extreme lighting effects, or shadow-related bugs inherent to the CARLA simulator.
\end{itemize}

\subsection*{DTP-Real}
DTP-Real was created using existing images from nuScenes (nuScenes v1.0 trainval split), adding only the annotations. The high-level pipeline followed to create the samples of DTP-Real is described in Algorithm \ref{algorithm_2}.

\begin{algorithm}
    \caption{The creation pipeline of DTP-Real}
    \label{algorithm_2}
    \begin{enumerate}
        \item Load nuScenes API.
        \item Using the annotations provided by nuScenes \\ identify which images are adequate for the type of question at hand.
        \item Create the corresponding annotations associated \\ with each image (e.g. filename, question, answer, distance, etc.)
        \item Save the annotations.
    \end{enumerate}
\end{algorithm}

\begin{figure}    
\centerline{\includegraphics[width=\columnwidth]{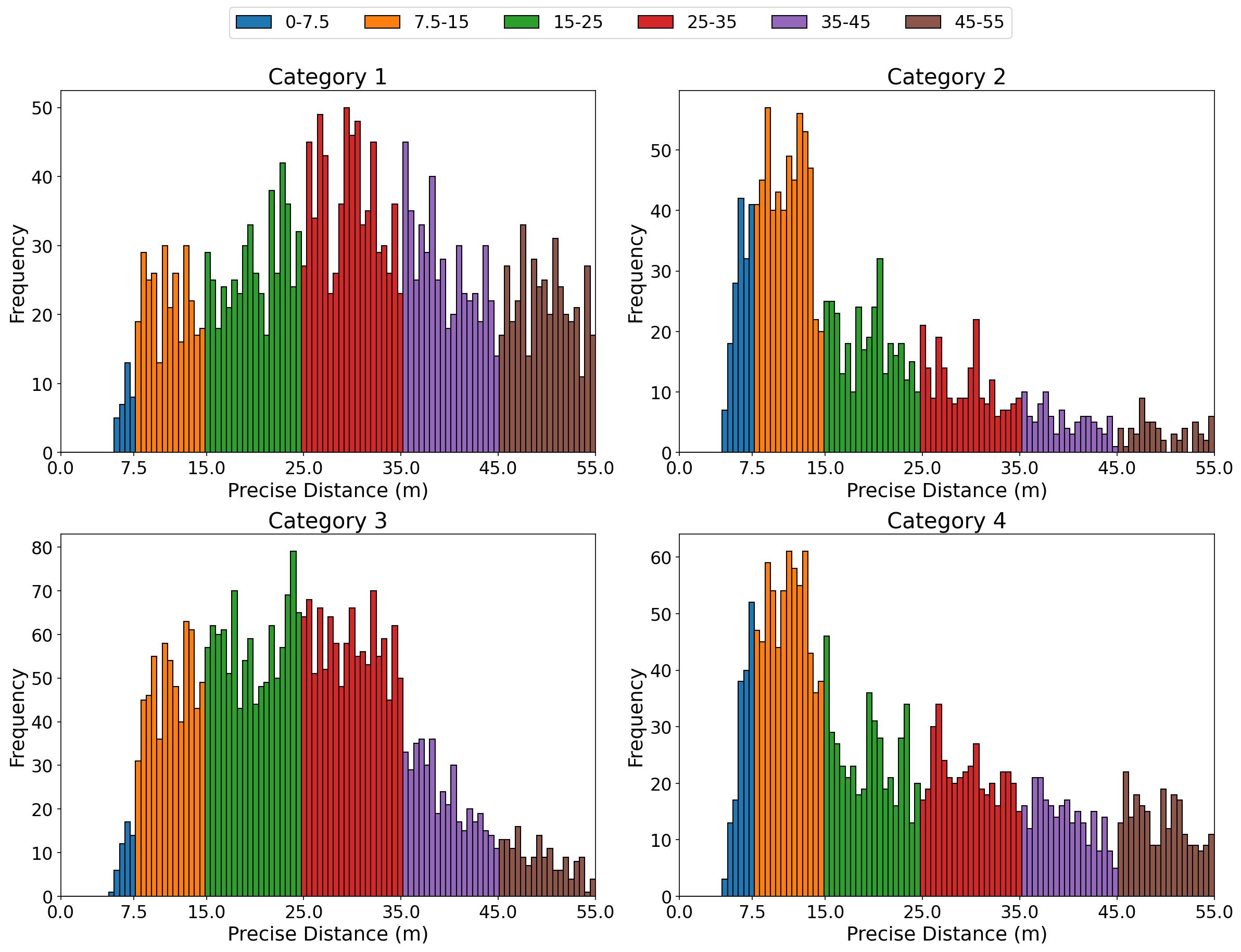}}
    \caption{Precise distance distribution in DTP-Real. \label{dtp_real_distance_distribution}}
\end{figure}

Regarding distance, unlike in the synthetic counterpart, we could not place the object in question at an exact desired distance. Instead, we created distance bins corresponding to 5, 10, 20, 30, 40, and 50 meters, and assigned each sample to the nearest bin. However, we also retained the exact distance of the object in question in the annotations, in case someone wishes to use it. We discarded all samples in which the distance of the object/agent in question was larger than 55 meters. In Figure \ref{dtp_real_distance_distribution}, we show the distribution of the precise distances for each data category in DTP-Real.

It is also worth noting that for DTP-Real we use the word \textit{person/people} in the questions instead of \textit{pedestrian(s)}, as there are images that depict people in the scene who are not pedestrians, such as construction workers, police officers, etc. Below, we outline the details of creating the samples for each category.

\subsubsection*{Category 1 (Person in the scene)}
For Cat.1-Real, we needed images containing exactly one person in the scene for the positive samples (answer = ``Yes") and images with no people for the negative samples (answer = ``No"). We achieved this using the annotations provided by nuScenes. More specifically, we first set aside all keyframes containing at least one human (i.e., human.pedestrian.adult) in the sample annotations. From these, we selected those with exactly one human who was fully visible (i.e., visibility level 4) from the chosen camera (the front camera in our case). We then saved these selected samples along with the necessary annotations, such as filename, the corresponding question and answer, and the distance of the object in question (both exact and binned). This resulted in the 2,381 positive samples listed in Table \ref{dtp_detailed_stats} for Cat.1-Real. Finally, we collected all keyframes without humans in their annotations and randomly selected 200 of them (again from the front camera) for inclusion in DTP-Real.

\subsubsection*{Category 2 (Person direction)}
The data collection for Cat.2-Real was similar to Cat.1-Real, with two additional requirements. First, we only wanted images showing people walking, not standing (so that the question about the direction makes sense). Second, even if the person was walking, we required a specific direction so that the answer would be \textit{left} or \textit{right}.

After collecting all images with exactly one person, using the same process as for Cat.1-Real\footnote{For Cat.2-Real, we used images from all six surround view cameras of nuScenes (i.e. front camera, front-right camera, front-left camera, back camera, back-right camera, back-left camera), unlike Cat.1-Real, because the additional constraints would otherwise result in very few samples.}, we used the velocity annotations from nuScenes to exclude samples where the person had a velocity below 0.5 m/s. We then used the rotation annotations of the camera, ego vehicle, and human to calculate the human’s angle relative to the camera. We kept only images where the human’s direction was clearly to the left or right from the camera’s point of view, discarding the rest. We defined \textit{left} as an angle within $90^\circ \pm 20^\circ$ and \textit{right} as an angle within $-90^\circ \pm 20^\circ$ from the camera's point of view.

\subsubsection*{Category 3 (Multiple people in the scene)}
The process for Cat.3-Real was similar to Cat.1-Real, but this time we repeated the procedure for images containing exactly two and exactly three people. We excluded images with exactly four people, as they were very rare and would reduce the total number of samples in this category, since we aim for a balanced number of samples for each answer.

Since there are multiple objects of interest in this case and we cannot place them close together as in the synthetic counterpart, each person in the scene is at a different distance from the camera. We define the characteristic distance of a sample as the average distance of all people in the scene. However, we also store the individual distances of each person and the variance of these distances in the annotations, as these may be of interest.

For samples with zero or one person in the scene, we reused some of the data collected for Cat.1-Real. The only new samples were those with two or three people, which we collected from all six cameras due to their lower frequency.

\subsubsection*{Category 4 (Person's location)}
For Cat.4-Real, we again used only keyframes with a single person in the scene to avoid ambiguities. The only additional step compared to Cat.1-Real was determining the type of surface the person was standing on. To do this, we used the pedestrian’s translation annotation (i.e., their coordinates on the map) and the map information, which specifies the surface type at each coordinate. We kept only samples where the person was on a road, sidewalk, or crosswalk, discarding the rest. Here, too, we used images from all six cameras to obtain a sufficient number of samples.

\subsubsection*{Data balancing}
The final step was to balance the number of samples corresponding to each answer for each category in DTP-Real. For example, in Cat.4-Real, the number of samples for ``road", ``sidewalk", and ``crosswalk" differed. When evaluating \glspl{vlm}, we do not want model $A$ to have an advantage over model $B$ simply because it is biased toward answering ``sidewalk" and the dataset happens to contain more such samples.

To avoid this, we ensured that each answer appeared the same number of times by identifying the answer with the fewest samples for each category and randomly removing samples from the others to match this number. For categories 1 and 3, this balancing was applied to the positive answers only, excluding the negative samples.

\section*{VALIDATION AND QUALITY} 
DTP-Synthetic was created using CARLA \cite{Dosovitskiy2017}, one of the most widely used simulators in the autonomous driving community. Although no simulation is perfect, CARLA is generally regarded as one of the best available options for such work \cite{kaur2021survey}. Moreover, since all simulators occasionally produce errors, we manually inspected all images in DTP-Synthetic and removed any problematic samples, as already discussed in \hyperref[collection_methods]{COLLECTION METHODS AND DESIGN}. This process ensures the high quality of DTP-Synthetic.

DTP-Real, on the other hand, was built on top of nuScenes \cite{Caesar2020}, one of the most widely used datasets in the autonomous driving community. Because DTP-Real relies exclusively on the images and annotations provided by nuScenes, it inherits the quality standards of nuScenes, which, although not flawless, is considered a high-quality dataset \cite{liu2024survey}.

\begin{figure}              \centerline{\includegraphics[width=3.5in]{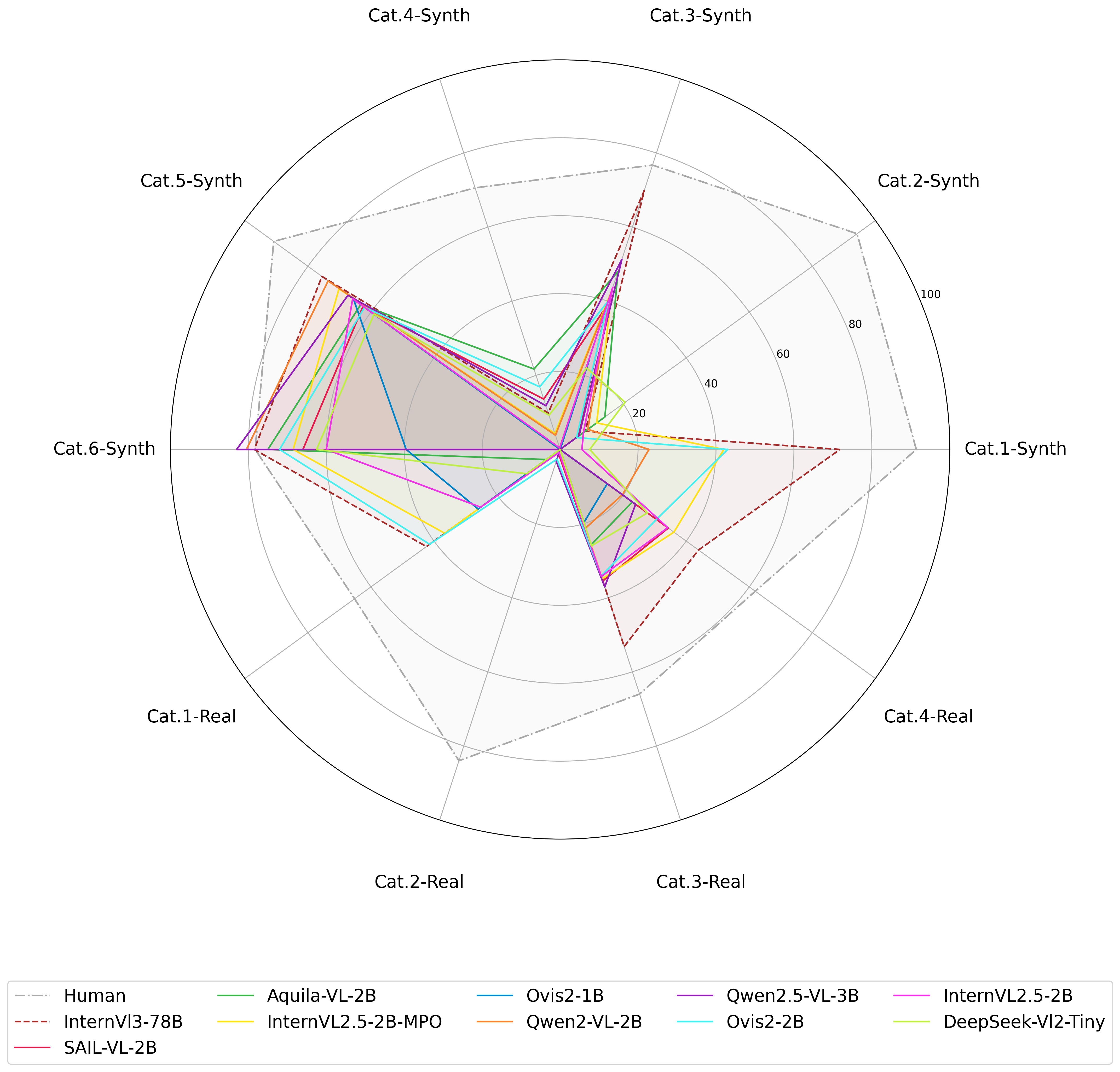}}
    \caption{Chance-corrected accuracy (\%) on \gls{dtpqa} \cite{11230063}. \label{normalized_radar_chart}}
\end{figure}

Finally, the large gap between human and model performance further demonstrates the high quality of \gls{dtpqa}. As shown in \cite{11230063}, \gls{sota} small \glspl{vlm} perform significantly worse than humans, particularly on tasks that require spatial perception (e.g., Cat.2-Synth, Cat.4-Synth, Cat.2-Real). Figure \ref{normalized_radar_chart} illustrates this performance gap, displaying the chance-corrected accuracy of nine \gls{sota} small \glspl{vlm} (and one big \gls{vlm}) compared to human accuracy across \gls{dtpqa} tasks. This substantial discrepancy indicates that \gls{dtpqa} effectively challenges current models and serves as a valuable benchmark for evaluating the perception capabilities of \glspl{vlm} in traffic scenarios.

\section*{RECORDS AND STORAGE} 
The structure of the main directory of \gls{dtpqa} is shown in Figure \ref{dirtree}. The dataset contains two types of files: the DTP-Synth images (.jpg files) and the annotations file (.json file), which contains annotations for both DTP-Synth and DTP-Real. It is important to note that the provided package is only partially self-contained. Specifically, the DTP-Real images are not included and must be downloaded separately as part of the nuScenes v1.0 trainval split\footnote{nuScenes data can be downloaded \href{https://www.nuscenes.org/nuscenes}{here}.}. The annotations file is a Python dictionary saved in JSON format. Its structure is depicted in Listing \ref{annotations}.

\begin{figure}[h]
    \dirtree{%
    .1 root/.
    .2 dtp\_synth/.
    .3 pedestrian\_crossing/.
    .4 1738940130087.jpg.
    .4 1738940131879.jpg.
    .4 ....
    .3 multiple\_pedestrians\_crossing/.
    .4 1739202245107.jpg.
    .4 1739202247458.jpg.
    .4 ....
    .3 blinker/.
    .4 1739964613847.jpg.
    .4 1739964626571.jpg.
    .4 ....
    .3 traffic\_lights/.
    .4 1740412762177.jpg.
    .4 1740412765106.jpg.
    .4 ....
    .3 traffic\_signs/.
    .4 1740579245570.jpg.
    .4 1740579248548.jpg.
    .4 ....
    .2 annotations.json.
    }
    \caption{Directory structure of the DTPQA dataset.}
    \label{dirtree}
\end{figure}

\begin{lstlisting}[
    float,
    language=Python,
    caption={Detailed JSON schema and sample values for the annotations.json file in DTPQA. Note: $^*$ indicates that the specific field is included only for Cat.5 and Cat.6 samples, while $^{**}$ is included only for Cat.3.},
    label={annotations},
    showstringspaces=false,
    mathescape=true
]
{
  "synth": {
    "category":
      {
        "image_path" (str): "...1738940130087.jpg"
        "question" (str): "Are there any...?"
        "answer" (str): "Yes"
        "distance" (int/None): 50
        "precise_distance"$^*$ (float/None):  48.578
        "town" (str): "Town01"
        "weather" (str): "Wet Cloudy Sunset"
      }
    ]
  },
  "real": {
    "category":
      {
        "image_path" (str): "...1884002262460.jpg"
        "question" (str): "How many people...?"
        "answer" (str): "Two"
        "distance" (int/None): 10
        "precise_distance" (float/None): 11.763
        "precise_distance_1"$^*$$^*$ (float/None): 7.587
        "precise_distance_2"$^*$$^*$ (float/None): 15.939
        "precise_distance_3"$^*$$^*$ (float/None): None
        "variance"$^*$$^*$ (float/None): 8.903
      }
  }
}
\end{lstlisting}

The annotations for each synthetic sample include the image path (relative to the root directory of \gls{dtpqa}), the corresponding question and answer, the distance of the object/agent from the camera, and the town and weather settings used in CARLA to generate the image. Cat.5-Synth and Cat.6-Synth also include the precise distance of the object, as this can differ slightly from the distance bin, as explained in \hyperref[collection_methods]{COLLECTION METHODS AND DESIGN}.

The annotations for each real sample include the image path, the corresponding question and answer, the distance bin (under the key \texttt{distance}) and the precise distance of the agent from the camera. The image path is specified relative to the nuScenes root directory. Therefore, once the nuScenes images are downloaded, the DTP-Real annotations can be used directly by prepending the root directory path to the image path. Additionally, Cat.3-Real includes the precise distances of all individuals in the scene, as well as the variance of their distances in cases where multiple people are present.

\section*{INSIGHTS AND NOTES}
It is important to note that \gls{dtpqa} is a \gls{vqa} dataset for evaluating the perception capabilities of \glspl{vlm} in simple traffic visual questions as a function of the distance of the object/agent in question, and this is how it was used in our study mentioned in \hyperref[background]{BACKGROUND}. However, it can also be used in slightly different ways. First of all, \gls{dtpqa} can be used to study how the performance of \glspl{vlm} on simple visual tasks is affected by small changes in the phrasing of the question or the structure of the prompt in general, as it is easy to replace the existing questions with similar ones when running the experiments. We investigated this direction to some degree in our study \cite{11230063}, but more detailed research is needed.

In addition, \gls{dtpqa} includes extra annotations that could support different types of studies. For example, in DTP-Synth we provide weather annotations, which would allow for a weather-dependent study of the perception capabilities of \glspl{vlm}. In Cat.3-Real we include the variance of the distances of people in the scene when multiple people are present. This could enable a variance-dependent study of the counting capabilities of \glspl{vlm} at a constant distance. The town annotations offered in DTP-Synth could also be useful for various analyses. Finally, \gls{dtpqa} can also be used for fine-tuning \glspl{vlm}, or even for training as part of a larger dataset, which are directions we have not yet explored.

\section*{SOURCE CODE AND SCRIPTS}
DTP-Synthetic was created using CARLA version 0.9.15\cite{Dosovitskiy2017} and DTP-Real was created on top of nuScenes v1.0 trainval split \cite{Caesar2020}.
The code used to create \gls{dtpqa} can be found in the corresponding repository on GitHub \cite{dtpqa}, along with instructions for generating DTP-Synthetic and DTP-Real\footnote{DTP-Real requires prior download of the nuScenes v1.0 trainval split.} in the respective README.md files.

\section*{ACKNOWLEDGEMENTS AND INTERESTS}
This dataset is partially based on the nuScenes dataset by Motional. We thank the creators of nuScenes for creating the original dataset, which we extend by introducing additional annotations.
N.T. created the dataset. C.E. acquired the necessary funding. C.E. and A.S. conceived and supervised the project. N.T. wrote the initial draft of the paper, and all the authors contributed to revisions.  
The authors have declared no conflicts of interest.

\bibliographystyle{IEEEtran}
\bibliography{library}

\end{document}